%% file: main.tex
\documentclass[11pt,a4paper]{article}
\usepackage[hyperref]{emnlp-ijcnlp-2019}
\usepackage{times}
\usepackage{amsfonts}
\usepackage{latexsym}
\usepackage{graphicx}
\usepackage{url}
\usepackage{algorithm}
\usepackage{algpseudocode}
\usepackage{diagbox}
\usepackage{amsmath}

\aclfinalcopy

\title{Detecting and Reducing Bias in a High Stakes Domain}

\author{ 
  Ruiqi Zhong\textsuperscript{1} \thanks{The first author is now a PhD student at University of California, Berkeley.}, 
  Yanda Chen\textsuperscript{1},
  Desmond Patton\textsuperscript{2},
  Charlotte Selous\textsuperscript{2}, 
  Kathleen McKeown\textsuperscript{1}
  \\
  \textsuperscript{1}Department of Computer Science, Columbia University \\
  \textsuperscript{2}School of Social Work, Columbia University\\\AND
  {\tt \{rz2383, yc3384, dp2787, cis2114\}@columbia.edu} \\
  {\tt \{kathy\}@cs.columbia.edu}
  }

\date{}

\begin{document}
\maketitle
\input{Abstract.tex}
\input{Intro.tex}
\input{Discover.tex}
\input{Data.tex}

\input{Methods.tex}
\input{Metrics.tex}
\input{Results.tex}
\input{Related.tex}

\input{Discussions.tex}

\bibliography{emnlp-ijcnlp-2019}
\bibliographystyle{acl_natbib}
\clearpage
\appendix
\input{appendix.tex}

\end{document}

%% file: Abstract.tex
\begin{abstract}
  Gang-involved youth in cities such as Chicago sometimes post on social media to express their aggression towards rival gangs and 
  previous research has demonstrated that a deep learning approach can predict aggression and loss in posts.
  To address the possibility of bias in this sensitive application, we developed an approach to systematically interpret the state of the art model. 
  We found, surprisingly, that it frequently bases its predictions on stop words such as ``a'' or ``on'', an approach that could harm social media users who have no aggressive intentions. 
  To tackle this bias, domain experts annotated the rationales, highlighting words that explain why a tweet is labeled as ``aggression". 
  These new annotations enable us to quantitatively measure how justified the model predictions are, and build models that drastically reduce bias. 
  Our study shows that in high stake scenarios, accuracy alone cannot guarantee a good system and we need new evaluation methods. 
\end{abstract}

%% file: Intro.tex
\section{Introduction}
There has been increased interest in using natural language processing to help understand the root causes of violence in cities such as Chicago. Researchers in social work found that posts on social media by gang-involved youth mirror their life on the physical street~\cite{patton2016sticks, patton2017tweets} and suggested that if community outreach workers could use tools to help identify posts of aggression and grief, they could intervene before grief turns to retribution and help avert violence. Early research in this area~\cite{blevins2016automatically} used support vector machines for the three-way classification task of {\em aggression}, {\em loss} and {\em other} and recent work~\cite{Chang2018DetectingGE} demonstrated a significant increase in accuracy through the use of a neural net approach along with the representation of contextual information from previous conversations. In both of these approaches, computer scientists and social work researchers worked side by side to enable qualitative analysis of the data to use in machine learning and to provide meaningful error analyses. 

Given the interest in using this work in a live setting to identify situations where community outreach workers could intervene, we are concerned with 
\textit{how}
the model makes the decision.
This is especially important since this is a high stakes domain and machine learning systems are known to incorporate bias. 
For example, the COMPAS Recidivism Algorithm, which assesses the likelihood of a criminal defendant to re-offend and influences judges' sentencing decisions, is known to be biased against African Americans~\cite{feller2016computer}. 
To avoid similar errors in the gang-violence application, 
we examined whether the model is biased through systematic investigation of how the model makes the predictions. 
Since misclassifying a tweet as \textit{aggression} might  potentially criminalize a vulnerable population (young black and latino youth) that is already over criminalized, we focus on classification of the \textit{aggression} class in our paper.

Our finding is surprising: in at least $10\%$ of the cases the \citet{Chang2018DetectingGE} model relies on stop words such as ``a'' and ``on'' to classify a tweet as aggression; these words carry no semantic or pragmatic indication of aggression or loss.  
We recognized that there might be more hidden biases, so we developed novel metrics to quantitatively compare the difference between how humans and models make their classification.
To achieve this, we ask domain experts to annotate  
the rationale for the aggression tweets by highlighting words and phrases that indicate why  experts classify them as aggression. We  compare the rationale against the models' interpretation. 

We also develop a model to incorporate human rationales, and  show that the system can make predictions based on similar reasons to those of domain experts. 
Our F-score is slightly improved over \citet{Chang2018DetectingGE} and we are able to reduce the bias drastically. We show how important this is by using an adversarial approach which demonstrates that we can change the prediction of posts predicted as {\em other} to {\em aggression} by the simple insertion of the non-offending function words. 
This implies that high F-score alone does not guarantee a good system, and that researchers in other high stakes domain should evaluate their algorithms from various angles.

Our contributions include:
\begin{itemize}
    \item A validated finding that the previous state-of-the-art model is biased.
    \item A rich dataset created by domain experts containing annotated human rationales.
    \item New metrics to evaluate models' robustness under adversary, and the difference between model interpretations and human rationales.
    \item A model that incorporates human rationales in training and behaves more as humans do. 
\end{itemize}

We also include an in-depth discussion on how misinterpretation might emerge when labels are being annotated by the experts. 
The battle against bias is far from solved, and we hope our insights and methodologies can guide future research in high stakes domains. 


%% file: Discover.tex
\section{Revealing Model Biases}

We used the code and data from \citet{Chang2018DetectingGE} and fully reproduced the state-of-the-art model that uses a Convolutional Neural Network, domain specific word embeddings and context features. 

\subsection{Locating Biases} \label{locate}
We examine the model using the simple leave-one-out method similar to \citet{ribeiro2016should}.
Suppose each tweet $T$ is a sequence of $l$ tokens $[e_{1}, e_{2} ... e_{l}]$ and the model gives the prediction confidence score $y$. 
For each token $e_{i}$, we calculate the model's prediction confidence $y_{-i}$ on the same tweet with $e_{i}$ masked as an unknown word;
hence the influence of the $i^{th}$ token $I_{i}$ is approximated by $I_{i} = y - y_{-i}$.
Then we manually examine the top two influential uni-grams for each tweet that the model classified as aggression.

As we expect, tokens strongly associated with aggression, such as ``\includegraphics[height=.8em]{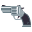}", ``dead" and many other curse words appear frequently as the most influential tokens.
However, surprisingly, words such as ``a'' and ``on'' are frequently considered the most influential by the model as well.
Out of 419 tweets that the model classifies as aggression, 15/32 times the model considers the token ``a'' to be the top
first or second most influential token and ``on'' 24/51 times the top first or second, respectively.
We also found that the model is biased against other stop words, such as ``into" and ``of".
This indicates that more than \textbf{10\%} of the time the model heavily relies on these unintentional differences of language use to classify tweets as aggression!

\subsection{Confirming Biases Exist}
Initially, we found this unexpected result 
surprising, so we checked the second order sensitivity \cite{feng2018pathologies} and constructed adversarial data sets. 

\textbf{Second Order Consistency} 
\citet{feng2018pathologies} defined this term when they found that the leave-one(token)-out (LOO) method might find a completely different set of most influential words if the most un-influential word is dropped from the sentence.
However, we do not observe this problem in our case: out of 419 predictions, 379 times the most influential word remains the most influential and 33 times it becomes the second most influential one after we drop the most un-influential words. 

\textbf{Adversarial Dataset}
If we can identify a weakness of the model, we must be able to break it in a corresponding way. 
To manifest that there is indeed bias against ``a'' and ``on,''  for each of these two words, we inserted it in each labeled tweet and counted the number of times the model changes its prediction from non-aggression to aggression. 

To ensure that our insertions are natural, we trained a large language model \cite{peters2018deep} on the unlabeled corpus ($\sim 1$ million tweets) used by \citet{Chang2018DetectingGE}, and score the naturalness of the insertion by the change of perplexity given by the language model.
For each tweet we calculate the most ``natural" position to insert our biased word and only 
took the top 800 natural tweets out of $\sim$ 7000 candidates as our adversarial dataset. 
We also subsample this dataset and check the subsample manually to make sure that these insertions make sense. 
In fact, since most tweets in the dataset are grammatically flexible in the first place, our insertion methods usually do not make them more unnatural - we have fewer constraints than adversarial attacks \cite{alzantot2018generating} conducted in domains such as IMDB reviews \cite{maas-EtAl:2011:ACL-HLT2011} or Textual Entailment \cite{bowman2015large}.

We find that even with these naive and seemingly harmless insertions, the model changes the prediction from non-aggression to aggression for about 3\%/5\% of the generated tweets when inserting ``a"/``on". 
With this overwhelming evidence, we conclude that the model is biased. 

The simplest way to fix this specific bias is a post-mortem one - just remove ``a'' and ``on'' in training and prediction. 
However, this only works if these are the only two points of bias and 
we suspect that there are many more biases lurking in the model. Hence, we need to comprehensively study the difference between how humans and models make their decision.
We need a new dataset with annotated human rationales to achieve this goal.

%% file: Data.tex
\section{Data} \label{data}
We use the same dataset and cross validation folds as in \citet{Chang2018DetectingGE}, which contains 329 Aggression tweets, 734 Loss tweets, with the Other class comprising the remaining 3,873 tweets.
The language in the dataset is hyper-local to the community we are studying and differs significantly from standard American English.
Therefore, most of the toolkits (e.g. pre-trained language models, parsers, etc) developed for standard English do not apply.
%
We extend their dataset by asking domain experts to annotate the rationales and additional information on context for each of the tweets labeled as aggression.
Below is the set of annotation instructions. 
\begin{itemize}
    \item \textbf{Rationales in Tweets} 
  We ask the domain experts to annotate  the most important word/words (rationales) in the tweet that makes them think the tweet is aggression/loss; or say, the removal of which would make them not consider it as an aggressive tweet. 
    We ask them to annotate the ``minimum set of words" that is important for their final decision, thus excluding words that are not very meaningful, such as ``on", ``basically", ``lol", etc. 
    Examples are shown in Table \ref{tab:examples}. 
    \item \textbf{Aggressive by Context}
    Sometimes the tweet itself does not contain aggressive content if we interpret them at face level, but placing it into context reveals that the user in fact conveyed aggressive intent.
    We ask the experts to label this attribute as True whenever they use an outside source to provide additional information needed to label a tweet. 
    These sources can include news sources, past posts, interactions, online dictionaries, or reddit threads.
    
    \item \textbf{Aggressive by Mention/Retweet} 
    Similarly, interactions between users that are
    suspected to be affiliated with the same or rival gangs might contain aggressive intent. For example, users may communicate with others in the same gang about the possibility of retaliation or they may directly taunt members of rival gangs.
    We ask the annotators to label this attribute as True if the context of interaction between users contributes to the aggression label.
    This corresponds to the pairwise interaction feature in \citet{Chang2018DetectingGE}.
    \item \textbf{Aggressive by Picture/URL} This attribute indicates whether the picture or URL in the tweet contains aggressive content.
    Since our study is natural language processing based, we do not make use of this piece of information in this paper.
    \item \textbf{Controversial Labels} Occasionally our experts would interpret a tweet previously considered aggressive as non-aggressive.
    We do not use these labels in our experiments for fair comparison with \citet{Chang2018DetectingGE}, but we include them in our new dataset.
\end{itemize}
    
\def\bt#1{\boxed{\text{#1}}}
\begin{table}
  \centering
  \begin{tabular}{l}\hline
  \textbf{Tweets} \\ \hline ``@user\_ do you think it's cool yo ass\\ slow ? You sound \underline{dumb} \underline{asf}'' \\ ``Rob you fucking \underline{goofys} that got some to say\\ about \underline{rage} just to show y'all bitches'' \\ ``GlizzyGang Bitch We Got Out \underline{Glocks} Up\underline{\includegraphics[height=.8em]{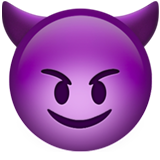}}\\ \hline \hline
  \end{tabular}
  \caption{Typical aggression tweets in our dataset, which includes annotations with human rationales (underlined).}
  \label{tab:examples}
\end{table}

Out of the 329 aggression tweets, 274 contain rationales in the tweet, while the other aggression labels are inferred by context rather than the content within the tweet itself. 
To keep notations uncluttered, we refer to all the ``aggressive by *" labels as \textit{context labels} in the rest of this paper.
The meta-statistics on this can be seen in table \ref{tab:by-meta}.
Additionally, 39 (12\% of aggression) tweets are considered controversial by our experts. 
We also ask the annotator to write down thoughts and observations while annotating, and we
share these insights in our discussion section. 
\footnote{Due to privacy concerns we cannot release the additional annotation publicly, but we will make it available to researchers who sign an MOU specifying their intended use of the data and their agreement with our ethical guidelines. Send email to the dp2787@columbia.edu.
Please contact the authors of \citet{Chang2018DetectingGE} for the original dataset.}

\begin{table}[h]
\centering
\begin{tabular}{|c|c|c|c|} 
\hline 
Context Label Types & Yes & No & N/A\\ \hline
Context & 107 & 74 & 148\\ \hline
Mention/Retweet & 36 & 35 & 258\\ \hline
URL & 9 & 3 & 317\\ \hline
Picture & 14 & 15 & 300 \\ \hline
\end{tabular}
\caption{Whether aggressive intent is observed in Context, Mention/Retweet, URL and/or Picture. Many of the context label cannot be annotated due to Twitter access policy.}
\label{tab:by-meta}
\end{table}

%% file: Methods.tex
\section{Methods}
We compare the Convolutional Neural Network (CNN) model from \citet{Chang2018DetectingGE}, an LSTM (Long Short Term Memory) Network model with attention mechanism (architecture defined below) and our new model that incorporates rationales by trained attention \cite{zhong2019fine}.\footnote{ Our code is publicly available at \url{https://github.com/David3384/GI_2019}}

\subsection{Overview of \citet{Chang2018DetectingGE}}
Here we give a quick overview of their CNN model with context features.
Let $l$ denote the length of the pre-processed tweet, $T = [e_{1}, e_{2} ... e_{l}]$ be the input sequence, and $c$ the context feature. 
First each token $e_{i}$ is mapped to an embedded representation $w_{i}$  through their domain specific word embeddings.
Then they used a convolution layer with kernel size 1 and 2, followed by a maxpooling layer and a linear fully connected layer to produce the tweet representation $z$. 
Finally, they concatenated $z$ with the context feature $c$ and fed it to the final classification layer. 
We refer the readers to their original paper for further details.

\subsection{LSTM with Attention} \label{lstm_archi}
We use a standard LSTM model with attention mechanism.
The model encodes the embedded sequence $[w_{1}, w_{2}, ..., w_{l}]$ with an LSTM to obtain hidden state representation $h_{i}$ with hidden dimension 64 for each position $i$ and then calculate an attention weight for each position based on $h_{i}$. 
The final representation $z$ is a weighted average of the hidden states concatentated with the context feature, which we pass to the final classification layer.
Formally,
\begin{equation}
    \alpha_{i} = tanh(\mathbf{v}h_{i}), A(i) = \frac{e^{\alpha_{i}}}{\sum_{j}e^{\alpha_{j}}}
\end{equation}
\begin{equation}
    z = [\sum_{i=1}^{l}A(i)h_{i}, c]
\end{equation}
where $\mathbf{v}$ are trainable weights we used to calculate the attention. 
We used binary cross entropy loss between the final prediction and the ground truth label, which we denote as $\mathcal{L}_{clf}$

\subsection{Trained Attention with Rationale} \label{rationale}
 We guide the model attention to attend to human rationales.
 We introduce a ``ground truth attention" $A^{*}$, which is uniformly distributed over all the positions where the token is an annotated rationale.
 For example, if $R = \{r_{1}, r_{2} \dots r_{d}\}$ are indexes of the tokens that are rationale words, then the ground truth attention distribution $A^{*}$ is defined as:
 \begin{equation}
     A^{*}(i) = \frac{1}{d} \text{ if } i \in R, 0 \text{ otherwise}
 \end{equation}
 Then we impose a KL divergence loss component between the ground truth attention and the model's attention.
 We optimize the weighted sum between the label classification and the attention loss, i.e.
 \begin{equation}
     \mathcal{L}_{attn} = KL(A^{*}||A), \mathcal{L} = \mathcal{L}_{clf} + \lambda_{attn}\mathcal{L}_{attn}
 \end{equation}
where we fix $\lambda_{attn} = 4$ after tuning.

To make a fair comparison with \citet{Chang2018DetectingGE}, we use the same set of hyper-parameters and training procedures.
We find that taking the majority vote of an ensemble of 5 neural networks trained on the same data can improve the F-score by 1$\sim$2 points, so we report both the averaged and the majority vote result across 5 independent runs. 
We also experiment with representations from a large scale language model trained on our unlabeled corpus \cite{peters2018deep}, but it only brought marginal improvements. 
For simplicity we do not include these experiments in our paper.

%% file: Metrics.tex
\section{Evaluation Metrics}
Besides the macro F-score in \cite{Chang2018DetectingGE}, we use rationale rank to characterize the difference between how humans and models make the decisions. 
We also quantitatively evaluate how the model uses the context feature.
For fair comparison, we used the same labels as in \cite{Chang2018DetectingGE} even though some labels are re-labeled as ``controversial".

\subsection{Rationale Rank}
We first use the leave-one-out-method to approximate the influence $I_{i}$ of each token $e_{i}$ as documented in section \ref{locate}. 
If the model makes predictions based on similar rationale words as human, $I_{i}$ should be large when $e_{i}$ is a rationale word, and should be small when $e_{i}$ is not a rationale. 
Therefore, a rationale word should have lower ranks if all the tokens are sorted by its influence on model's prediction in descending order.
We define the influence rank of the $i^{th}$ token $rank(i)$ as the number of tokens in the tweet that have higher influence score than the $i^{th}$ token.

We further relax our requirement for the model: we do not have the unrealistic hope that the model is relying on every
rationale word; instead, we only consider the lowest ranking rationale. 
Formally, suppose there are multiple rationale words with indexes $R = \{r_{1}, r_{2} .. r_{d}\}$, 
then we define the ``rationale rank" as:
\begin{equation}
    rationale rank = min_{i\in R}rank(i)
\end{equation}
We calculate the average rationale rank for  all tweets that are classified as aggression and have rationale annotations on the cross validation test set for each fold. 

Since the majority of tweets usually has rationale rank 0, the averaged rationale rank is small and the difference between each model is ``diluted" by the prevailing zeros.
Therefore, for each model, we also report the percentage of tweets that has rationale 0 or 1, respectively.
This more intuitively reflects on what fraction of the tweets the models' reasoning agrees with human rationales. 


\subsection{Context Feature Evaluation} \label{context-metrics}
\citet{Chang2018DetectingGE} claimed that their pair-wise interaction feature captures information about users' past interactions, and they supported it by showing significant improvement when incorporating it into the CNN model.
Nevertheless, it does not automatically entail that the model is using this feature in a way that humans do. 
Here we examine this claim quantitatively with our annotated interpretation of context.

\citet{Chang2018DetectingGE} concatenated the context features before the final classification softmax layer, so we can directly calculate the impact of a context feature by taking its dot product with the corresponding weight learned by the model. 
Formally, let $c_{1}$ be the ``emotional state'' feature, $c_{2}$ the ``user history'', $c_{3}$ the ``pair-wise user interaction'' and the pooled tweet representation is $z$, then the final prediction score is given by:
\begin{equation}
    y = \sigma(w_{z}z + \sum_{i=1}^{3}w_{i}c_{i} + b)
\end{equation}
where, $w_{(\cdot)}$ are trainable weights and $b$ the bias term. 

We group the tweets by the annotated context label ``by mention/retweet'', a single label, annotated as described in section \ref{data}.
If the pair-wise interaction feature $c_{3}$ truly models domain experts' interpretation of user interaction as \citet{Chang2018DetectingGE} has hypothesized, $w_{3}c_{3}$ should be higher for those tweets that are annotated as ``aggressive by mention/retweet" than those that are not. 
We do not examine use of other context labels and features, since other context features do not model URL, pictures or a general type of context. 

\section{Prediction Flips under Adversary} \label{adv data}
\subsection{Generating Adversarial Tweets}
We consider the simplest operation: inserting certain candidate neutral unigrams： $u \in U = \{$ ``a", ``on", ``da", ``into", ``of", ``that"$\}$， that the model strongly associates with aggression.~\footnote{We describe how we choose these unigrams in the appendix \ref{other-uni}. We also experimented with ``in", ``be", ``do", ``any", ``u", ``out", ``basically", ``yea", ``ever", ``\includegraphics[height=.8em]{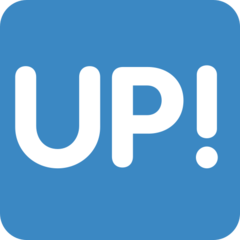}", ``\includegraphics[height=.8em]{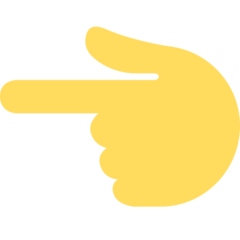}", ``we"， etc, but the results are not significant so we present them in the appendix.}
Assuming that we have already trained a language model that gives a perplexity score $p(T)$ for a tweet $T = [e_{1}, e_{2}, \dots e_{l}]$, 
then for each $u \in U$ we create a dataset by inserting it into all of the tweets $T$ in our labeled dataset $\mathcal{T}$.
We describe how we train the language model in the appendix section \ref{generate-advd}. 

For each $T \in \mathcal{T}$, there are $|T| + 1$ positions to insert the unigram $u$. 
We pick the resulting tweet $T'$ with $u$ inserted that has the highest likelihood score $p(T')$ among all the $|T| + 1$ possible insertions, and calculate the corresponding ``insertion naturalness score" $p(T') - p(T)$.
After inserting the unigram $u$ into all the tweets in the labeled dataset $\mathcal{T}$, we obtain a corresponding dataset $\mathcal{T}'$ and pick the top 800 tweets that have the highest ``insertion naturalness score". 

Then we calculate how many times a model would flip its prediction from non-aggression to aggression among these 800 pairs of tweets, before and after inserting the unigram $u$.
Notice that even one percent of ``flip rate" is significant: the unigrams $u$ we consider here are not rare words and thus have high coverage.
If the model is deployed at a large scale, millions of tweet will be classified, and tens of thousands of tweets will be influenced, not mentioning that we are considering only one unigram and there are many potential other tokens biased by the model.

\subsection{Evaluating Generation Naturalness}
Since words are discrete, generating adversarial text inputs has been a hard problem in other domains for at least two reasons: 
1) one perturbation might be large enough to change the label.
2) we cannot guarantee that the resulting text is still a natural input. 
However, these challenges do not play much of a role in our domain. 
First, the candidate words we are inserting are ``function words'', which we know should not cause the model to flip their predictions.
Second, the large proportion of tweets we are working with are grammatically flexible in the first place.
Therefore, in contrast to genres that follow the  conventions of Standard American English grammar, we suspect that adding function words will not make the tweets look unnatural to the domain experts and we test this hypothesis. 

To evaluate the plausibility of our generated tweets, we sample 18 tweets from all the generated tweets  and 18 existing tweets from our labeled dataset. 
Then we ask our experts to try their best to classify whether each tweet comes from model generation or the existing dataset.
We also ask them to annotate whether each tweet is ``natural", ``reasonable but awkward" or ``nonsensical".

The experts classified 25 of them correctly ($\sim 70\%$), only slightly better than the $50\%$ random guess baseline even though they are aware that we are generating tweets by inserting function words.
Out of 18 generated tweets, they considered 7 as ``natural", 8 ``reasonable but awkward" and 3 as ``non-sensical".
In comparison,  out of the 18 existing tweets, they considered 14 as ``natural'', 1 as ``reasonable but awkward'' and 3 as ``non-sensical.''
These results imply that although our generated tweets might not all be fluent, they are still 
reasonable enough as adversarial tweets. 
Interestingly, they also considered 3 existing tweets as ``non-sensical'', partly because the tweets are highly grammatically flexible and we told them that there are generated tweets. 

%% file: Results.tex
\begin{table*}[t]
\centering
\begin{tabular}{|c|c|c|c|c|c|c|}
\hline
Model & \vtop{\hbox{\strut Average}\hbox{\strut Macro F1}} & \vtop{\hbox{\strut Ensemble}\hbox{\strut Agg. F1}} & \vtop{\hbox{\strut Ensemble}\hbox{\strut Loss F1}} & \vtop{\hbox{\strut Ensemble}\hbox{\strut Other F1}} & \vtop{\hbox{\strut Ensemble}\hbox{\strut Macro F1}}\\
\hline
\citet{blevins2016automatically} & 63.5 & 34.5 & 67.3 & 88.7 & 63.5\\
\hline
\citet{Chang2018DetectingGE} & \textbf{68.2} & 41.3 & \textbf{75.4} & 91.3 & 69.4\\
\hline
CNN + Twitter & 65.8 & 36.2 & 73.9 & 91.3 & 67.1 \\
\hline
LSTM & 67.0 & 38.4 & 74.2 & 91.3 & 68.0 \\
\hline
LSTM + Rationale & \textbf{68.2} & \textbf{43.0} & 75.0 & \textbf{91.6} & \textbf{69.8} \\
\hline
\end{tabular}
\caption {Results comparing different model architectures. \citet{Chang2018DetectingGE} refers to the full CNN model with domain specific embedding and context feature; ``CNN Twitter" replaces the domain specific embedding with GloVe twitter embeddings \cite{pennington2014glove}; LSTM refers to LSTM Attention described in section \ref{lstm_archi}; LSTM Rationale refers to the same LSTM Attention with trained attention described in section \ref{rationale}.
For each column, the best performing entry is boldered.
}
\label{performances}
\end{table*}

\begin{table}[t]
\begin{tabular}{|c|c|c|c|c|c|c|}
\hline
Model & \vtop{\hbox{\strut Avg}\hbox{\strut RR}} &
\vtop{\hbox{\strut RR}\hbox{\strut = 0}} & 
\vtop{\hbox{\strut RR}\hbox{\strut = 1}}  \\
\citet{blevins2016automatically} & 1.70 & 0.60 & 0.13\\
\hline
\citet{Chang2018DetectingGE} & 1.42 & 0.54 & 0.17 \\
\hline
CNN + Twitter & 1.82 & 0.43 & 0.25 \\
\hline
LSTM & 1.73 & 0.50 & 0.15\\
\hline
LSTM + Rationale & \textbf{0.86} & \textbf{0.69} & 0.14\\
\hline
\end{tabular}
\caption {Results on (averaged) rationale rank (RR), fraction of tweets when RR = 0 and RR = 1, respectively. For (averaged) rationale rank and RR = 0, the best performing entry is boldered. 
We do not bolder the column RR = 1 since it is sub-optimal.
}
\label{rationale-performance}
\end{table}

\begin{table*}[t]
\centering
\begin{tabular}{|c|c|c|c|c|c|c|c|}
\hline
\backslashbox{Models}{Unigrams} & a & on & da & into & of & that\\
\hline
\citet{blevins2016automatically} & 44.16$^{!}$ & 121.00$^{!}$ & 37.44$^{!}$ & 117.72$^{!}$ & 36.84$^{!}$ & 74.48 $^{!}$\\
\hline
\citet{Chang2018DetectingGE} & 26.56 & 36.00 & 8.52$^{*}$  & 21.68 & 19.20 & 7.52$^{*}$\\
\hline
CNN + Twitter & 38.40 & 35.80 & 17.60 & 15.88 & 5.24$^{*}$ & 7.64\\
\hline
LSTM & 16.16 & 40.28 & 21.52 & 26.92 & 13.20 & 13.12\\
\hline
LSTM + Rationale & 13.52$^{*}$ & 23.16$^{*}$ & 12.40 & 10.28$^{*}$ & 11.40 & 8.96\\
\hline
\end{tabular}
\caption {The number of model's prediction flip from non-aggressive to aggressive, out of the 800 attacking tweets generated by inserting a specified ``neutral" unigram. 
To obtain a stable estimate, the count of number of flips of each model is averaged across 5 model runs and models trained on 5 folds - thus leading to non-integer results.
For each column, the worst performing entry is marked with ``!" and best with ``*". 
}
\label{adversarial_stats}
\end{table*}

\section{Results}
The model F-score performances are shown in table \ref{performances} and rationale rank in table \ref{rationale-performance}. 
We present our results for macro F-score,
Rationale Rank, Prediction Flips on Adversarial Dataset and Context Feature Evaluation. 
We also evaluate the models from \cite{blevins2016automatically} for completeness.
\footnote{For consistency, we used their model but first classify aggression vs. rest, rather than other vs. rest as in \cite{blevins2016automatically}. The training procedure is deterministic, so the ensemble gives the same performance as individual ones.}

\textbf{F-score} 
We first notice that taking the ensemble of 5 independent runs trained on the same data consistently improves the performance over averaged individual run performance, so we report all of our results on ensemble models.
Compared to \citet{Chang2018DetectingGE}, our LSTM+Rationale model differs by at most one point on all of the macro-F score metrics, and sometimes achieves the best performance. 

\textbf{Rationale Rank}
A closer look at the rationale rank result reveals a huge difference (p-value $< 10^{-5}$): our model significantly beats all other models on this metrics. 

\textbf{Prediction Flips under Adversary}
The results are shown in table \ref{adversarial_stats}. 
For each candidate unigram we count the number of times the model flips its prediction from non-aggression to aggression among the top 800 insertions.
By incorporating rationales into the model, we achieve the fewest flips for half of the unigrams and under none of the six cases does it flip the most predictions. 

These two results demonstrate that although the CNN model 
\citep{Chang2018DetectingGE} only slightly underperforms our model in terms of F-score performance, it behaves very differently behind the scene.
Our model is more robust and makes predictions that match human predictions.
This implies that traditional F-score measure alone neither provides us deeper understanding of the model, nor reveals the underlying biases. 

\textbf{Context Feature Evaluation}
We calculate the correlation between ``how the human uses context" and ``how the model uses context" as described in section \ref{context-metrics}.
The mean difference of $w_{3}c_{3}$ between the two groups split by the label value is 1.61 with statistical significance $10^{-3}$, indicating that the pairwise interaction feature correlates well with how domain experts interpret users' retweet/mentions.

%% file: Related.tex
\section{Related Work} \label{related}
\citet{patton2013internet} initiated a new field of study, coining the term {\em internet banging} to describe the phenomenon whereby online altercations lead to offline violence. 
This work is further motivated by challenging gun violence statistics in the city of Chicago which saw a 58\% spike between 2015 and 2016. 
In addition, the number of homicides stemming from physical altercations decreased during that period with little empirical evidence to explain this shift \cite{kapustin2016gun}. 
Investigating internet banging represents an important first step in understanding why gun deaths increase despite fewer physical altercations. 

\citet{blevins2016automatically} is a collaboaration between social work and computer science experts that brings together social work's substantive knowledge of communities, youth development, and gun violence with data science's innovative machine learning tools. 
They developed a new analytical system that leverages community participatory methods and machine learning to automatically detect the expression of loss and aggression on Twitter.
\citet{blevins2016automatically} collected and annotated tweets posted by Gakirah Barnes, a deceased gang members prolific on twitter, as well as her top communicators.
\citet{Chang2018DetectingGE} extended this dataset to be six times larger and achieved the state-of-the-art result by using a CNN.

Our interpretability method is  most similar to the leave-one-out (LOO) approach by \citet{ribeiro2016should}.
Nevertheless, it risks making the sentence ungrammatical, hence producing ``rubbish examples" \cite{goodfellow2014explaining} and pathological interpretations \cite{feng2018pathologies}. 
We explicitly address this concern by checking the second-order consistency proposed in \citet{feng2018pathologies}.
To further ensure that the removal of a word results in a ``blank state", erasure-based method proposed in \citet{li2016understanding} can be used. 
It would also be interesting to explore different interpretability methods, such as using the influence function \cite{koh2017understanding} or  nearest neighbor in the feature space \cite{strobelt2019s}.
Our method of automatically generating adversarial data according to a pre-identified weakness of a model without accessing the model output and parameters is most similar to \citet{jia2017adversarial}.
From \citet{alzantot2018generating}, we borrowed the idea of using a language model to ensure naturalness of generated sentences, and it is a promising next step to collect natural human-generated adversarial examples \cite{wallace2018trick}.
In the future, studying white-box \cite{ebrahimi2017hotflip, bvelohlavek2017using} and character level attacks \cite{grondahl2018all} are important directions to provide additional robustness in this critical domain.

Trained attention was explored in event detection \cite{liu2017exploiting}, machine translation \cite{mi2016supervised} and sentiment analysis \cite{zhong2019fine}. 
More sophisticated approaches include deriving attention from human rationales \cite{bao2018deriving}.
Finding better ways to make use of this kind of information is a promising avenue, especially in this low-resource domain. 

Ensuring machine learning fairness has increasingly caught researchers' attention \cite{corbett2018measure} since algorithms can create concerning negative societal impacts \cite{feller2016computer}. 
Biases might be based on gender \cite{bolukbasi2016man, zhao2017men}, race \cite{kiritchenko2018examining, rudinger2017social}, writing styles \cite{madnani2017building}, or other factors. 
More broadly, bias can also occur when annotators label the data. 
For example, \citet{golbeck2017large} found it hard to distinguish between mere offensive speech and hate speech.

%% file: Discussions.tex
\section{Discussion}
\subsection{Limitations of the Annotation Instruction}

The instructions for annotating rationales in tweets do not fully recognize the complex nature of language.
Current annotation reduces the experts' underlying reason to annotate a tweet as aggression to a set of key words, which is not flexible enough to convey more complicated inferences. 
The experts' analyses are far more complex: they use context to disambiguate the meaning of the rationale words rather than identifying key words.
Since the language is hyper-local and many of the non-aggressive tweets might appear aggressive in outsiders' eyes, interpreting the context is a crucial component while determining an aggression label and it is important to develop more flexible context rationale representations in the future.

Take, for example, the post ``I talk to alotta Da Guyz In jail Everyday On l'a Dese N***as Ain't Sending Bro Nem A Dime \includegraphics[height=.8em]{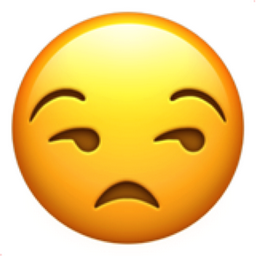} But Steady Yellin Free Em Goof ass…". 
At face value, the words "\includegraphics[height=.8em]{img/unamused-face.png}", ``goof", and ``ass" per se do not capture the reason why this post is aggressive;
the aggression label for this post in fact derives from the poster calling out the dishonesty of those who support incarcerated individuals in principle, without supporting them monetarily.

\subsection{Controversial Labels}
In the annotation procedure, it can be challenging to choose a single label when users express both anger and grief in a single post. 
In these instances, we aim to capture and annotate the tweet's dominant theme.

For example, the post ``Fuck Da System Fuck Da State Fuck 12z \includegraphics[height=.8em]{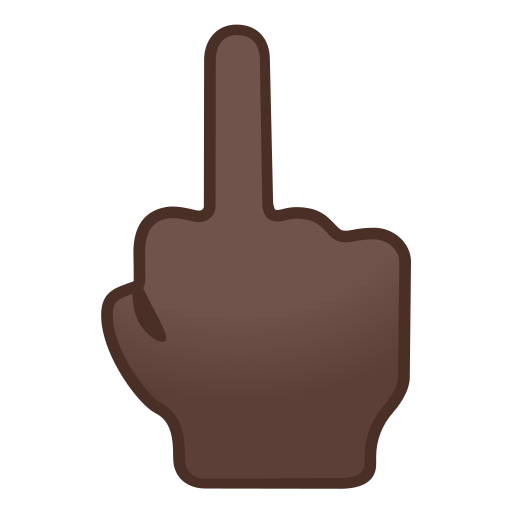}\includegraphics[height=.8em]{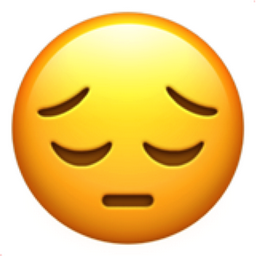}" contains elements of aggression--the word ``fuck" and the middle finger emoji--and loss--the sad frown emoji and the post's larger context. 
This post fits the \textit{aggression} label, by challenging authority and insulting, and the \textit{loss} label, by evoking incarceration. 
However, a \textit{loss} label is more suitable, because: 
1) though the anger seems to target specific players, the text as a whole reveals that the poster is highlighting institutional violence; and 
2) the sad frown emoji, paired with the context that one of the poster's loved ones was recently arrested on a drug charge ("12z" refers to law enforcement's drug unit), cements that the user is grieving in this post. 

\section{Conclusions and Future Directions}
In this paper, we interpret the previous state-of-the-art model, find it is biased against stop words, create a dataset of rationales, develop metrics to measure the difference between how models and humans make predictions, design an adversarial approach to test the robustness of the model, and propose a neural network architecture to incorporate rationales.
Our work has an important message to the broader NLP research community that develops tools for high stakes domains: the dominant evaluation methods in current research approaches, F-score and qualitative error analysis, are far from enough to develop good systems that might potentially lead to bad social consequences.
We need various different approaches to evaluate and build a model, and each of them requires further research for improvement:

\textit{Interpretability}: 
A solid understanding of model behavior can inform researchers of the potential weaknesses and guide the process of developing corresponding solutions. 
When trying to interpret \citet{Chang2018DetectingGE}'s deep learning model,  we find it biased against many function words.
Still, interpretation of neural network is an ongoing research topic; even though we verified ``second order consistency", future work is needed for better and more faithful explanations. 

\textit{Robustness under Adversary}:  
It is better to break the system before it is broken when it is deployed.
We find the previous model to be susceptible to insertions of certain function words, while our model is more robust against them.
Nevertheless, the model is not winning every column in table \ref{adversarial_stats}, so there is still room for improvement. 

\textit{Metrics Evaluating How Predictions are Made}: 
We need to quantitatively evaluate the difference between how models and humans classify tweets. 
We develop the metrics ``rationale rank" and ``prediction flips under adversary", and verify that our model behaves more similar to domain experts. 
F-score does not reflect this important improvement.
Additionally, uni-gram (lexical level) influence is only a shallow approximation of human reasoning - a process that involves understanding the semantics of a tweet as a whole and analyzing its pragmatics in the real-world context.
Ideally we hope to find a formal language to approximate such a process
and compare it against the model's interpretation. 

\textit{Models that Incorporate Rationales}: 
Our new model, though simple and straightforward, successfully makes use of the rationales and outperforms the previous model under most robustness metrics proposed in this paper; 
we believe that more sophisticated modeling of rationales can push the performances further, both in terms of accuracy and robustness.

\textit{Feature Evaluation}: 
Performance improvement caused by additional features does not entail that they are used by the model in an expected way.
We verify the claim of \citet{Chang2018DetectingGE} that the pairwise interaction feature does capture user retweet/mention interactions.
    
\textit{Domain Experts}: 
In our domain, stakes are high and user behavior is frequently subject to biases and misinterpretations.
Therefore, we ask domain experts rather than crowd workers to extend the annotation, and ask them to provide further insights about a domain that computer scientists are usually not at all familiar with. 
Furthermore, we should keep a critical eye on the labels and be aware of the limitations of our evaluation metrics for model accuracy. 
It is convenient for computer scientists to regard the labels as ground truth, the dataset as static and try different models to improve the f-score. 
However, as domain experts gained a deeper understanding about the twitter users, they marked 12\% of the aggression labels as controversial (section \ref{data});
the annotation may evolve over time.
Indeed, we need a consistent method and dataset to compare different models fairly, but we should also keep in mind that f-score is a proxy and might not necessarily reflect an improvement in the real-world.

%% file: appendix.tex
\section{Supplementary Materials}
\subsection{Training Language Model} \label{generate-advd}
We first describe the language model used to generate the adversarial dataset.
The language model is trained by the standard perplexity objective function, i.e., for a tweet $T = [e_{1}, e_{2} \dots e_{l}]$, we want the model to predict the next token $e_{i}$ based on the previous tokens $[e_{1}, e_{2} \dots e_{i-1}]$.
We train a language model for the forward and backward direction, respectively. 

As for the model architecture, we embed each token to create a real vector representation $[w_{1}, w_{2} \dots w_{l}]$;
then we stack three layers of single direction LSTM with hidden dimension 128 on top of the embedded token representation.
Then the hidden representation from the last LSTM layer is fed the final classification layer with a softmax activation, the output of which is the probability of each token in the entire vocabulary.

We randomly sampled $80\%$ of the tweets from the unlabled corpus ($\sim 1$ million tweets) used in \citet{Chang2018DetectingGE} as our training data, and use the rest as validation data.
We apply early stopping on our training by calculating the validation loss on the validation set. 

\subsection{Results for Other Unigrams} \label{other-uni}
We ran a simple $\ell_{1}$ logistics linear regression with unigram features on the aggression label to obtain a list of  unigrams that are highly correlated with the \textit{aggression} label.
This results in $\sim 300$ unigrams with positive weights, and we manually selected a subset of them based on two criteria:
1) possible to insert them into a significant proportion of tweets 
2) the experts do not use them to determine the label. 
Here we list the unigrams and their corresponding weight, and use bold front for selected unigrams. We report the adversarial flip results for other unigrams we have selected in table \ref{additional_adversarial_stats}.

\begin{table}[h]
\centering
\begin{tabular}{|c|c|} 
\hline 
Unigram & Corresponding Weight\\ \hline
\#lldv & 5.755\\ \hline
jj & 5.283\\ \hline
disrespect & 5.218\\ \hline
nobody's & 5.060\\ \hline
irritated & 4.968\\ \hline
keta & 4.864\\ \hline
delete & 4.791\\ \hline
boot & 4.700\\ \hline
snitchk & 4.698\\ \hline
saturday & 4.615\\ \hline
glock & 4.560\\ \hline
opps & 4.535\\ \hline
buses & 4.427\\ \hline
snitching & 4.418\\ \hline
sticks & 4.396\\ \hline
cpdk & 4.394\\ \hline
\includegraphics[height=.8em]{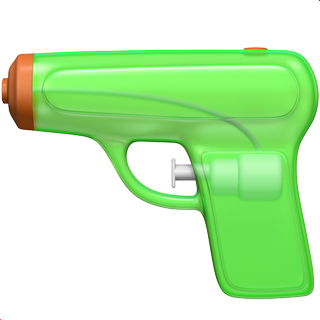} & 4.118\\ \hline
,bitch & 4.110\\ \hline
street & 4.057\\ \hline
situation & 3.881\\ \hline
guns & 3.675\\ \hline
snitchkkkkkkkkk & 3.662\\ \hline
ls & 3.659\\ \hline
rat & 3.581\\ \hline
nuskigang & 3.578\\ \hline
haters & 3.517\\ \hline
kill & 3.470\\ \hline
600 & 3.362\\ \hline
\includegraphics[height=.8em]{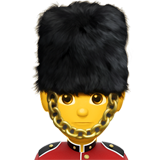} & 3.362\\ \hline
\#playcrazygang & 3.324\\ \hline
thinkin & 3.313\\ \hline
\includegraphics[height=.8em]{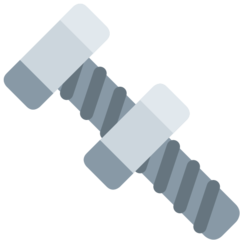} & 3.290\\ \hline
police & 3.284\\ \hline
mtg & 3.065\\ \hline
piss & 3.032\\ \hline
dumbass & 3.023\\ \hline
gatta & 2.972\\ \hline
pimp & 2.942\\ \hline
gunn & 2.843\\ \hline
\includegraphics[height=.8em]{img/up_exclamation.png} & 2.773\\ \hline
mitch & 2.772\\ \hline
bust & 2.748\\ \hline
poled & 2.690\\ \hline
dicks & 2.662\\ \hline
dot & 2.660\\ \hline
cype & 2.589\\ \hline
rayband & 2.463\\ \hline
\end{tabular}
\caption{Unigrams and their corresponding weights sorted in decreasing weight order}
\end{table}

\begin{table}[h]
\centering
\begin{tabular}{|c|c|} 
\hline 
Unigram & Corresponding Weight\\ \hline
.m & 2.448\\ \hline
boy & 2.364\\ \hline
blowing & 2.340\\ \hline
extorted & 2.292\\ \hline
gutta & 2.260\\ \hline
wacked & 2.244\\ \hline
guys & 2.238\\ \hline
wen & 2.230\\ \hline
lame & 2.164\\ \hline
blackin & 2.145\\ \hline
mfks & 2.138\\ \hline
\includegraphics[height=.8em]{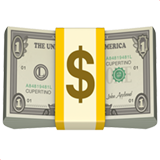} & 2.134\\ \hline
killas & 2.080\\ \hline
\includegraphics[height=.8em]{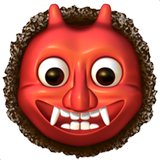} & 2.069\\ \hline
dead & 2.043\\ \hline
\includegraphics[height=.8em]{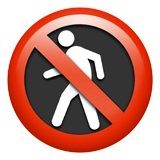} & 2.029\\ \hline
fuck & 2.022\\ \hline
30 & 1.963\\ \hline
n***as & 1.952\\ \hline
\includegraphics[height=.8em]{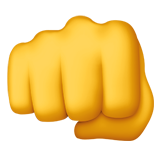} & 1.919\\ \hline
bih & 1.910\\ \hline
land & 1.900\\ \hline
thoe & 1.875\\ \hline
dthang & 1.852\\ \hline
800 & 1.851\\ \hline
bout & 1.810\\ \hline
strapped & 1.807\\ \hline
repping & 1.786\\ \hline
turn & 1.767\\ \hline
shoutout & 1.693\\ \hline
lucky & 1.680\\ \hline
\textbf{basically} & 1.674\\ \hline
ill & 1.669\\ \hline
mfs & 1.667\\ \hline
\includegraphics[height=.8em]{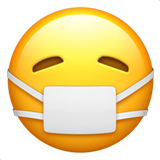} & 1.632\\ \hline
shut & 1.631\\ \hline
crease & 1.615\\ \hline
page & 1.603\\ \hline
shirt & 1.583\\ \hline
punching & 1.561\\ \hline
boa & 1.549\\ \hline
niqqas & 1.529\\ \hline
fight & 1.517\\ \hline
\includegraphics[height=.8em]{img/pointing_left.png} & 1.466\\ \hline
n***az & 1.463\\ \hline
riot & 1.453\\ \hline
ebt & 1.444\\ \hline
\end{tabular}
\caption{Unigrams and their corresponding weights sorted in decreasing weight order (continued)}
\end{table}

\begin{table}[h]
\centering
\begin{tabular}{|c|c|} 
\hline
Unigram & Corresponding Weight\\ \hline
fee & 1.438\\ \hline
blk & 1.437\\ \hline
pole & 1.405\\ \hline
o & 1.377\\ \hline
\includegraphics[height=.8em]{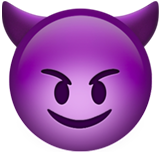} & 1.374\\ \hline
...lol & 1.369\\ \hline
hoes & 1.362\\ \hline
doer & 1.348\\ \hline
pussy & 1.347\\ \hline
wanna & 1.337\\ \hline
gotta & 1.331\\ \hline
another & 1.315\\ \hline
let & 1.303\\ \hline
mean & 1.297\\ \hline
watchem & 1.274\\ \hline
odee & 1.202\\ \hline
ready & 1.196\\ \hline
ass & 1.191\\ \hline
\textbf{into} & 1.175\\ \hline
@ & 1.113\\ \hline
put & 1.106\\ \hline
learn & 1.105\\ \hline
line & 1.100\\ \hline
\textbf{yea} & 1.087\\ \hline
\includegraphics[height=.8em]{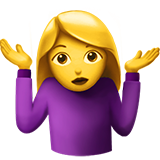} & 1.074\\ \hline
kids & 1.040\\ \hline
\includegraphics[height=.8em]{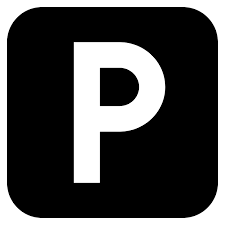} & 1.037\\ \hline
tell & 1.001\\ \hline
frontstreetsavagesquad & 0.981\\ \hline
got & 0.970\\ \hline
fucked & 0.962\\ \hline
run & 0.959\\ \hline
screaming & 0.944\\ \hline
changed & 0.940\\ \hline
delusional & 0.940\\ \hline
\#tyquanworld & 0.910\\ \hline
\textbf{any} & 0.909\\ \hline
talking & 0.878\\ \hline
wassup & 0.866\\ \hline
\includegraphics[height=.8em]{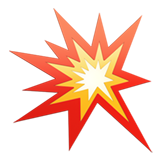} & 0.854\\ \hline
\textbf{on} & 0.846\\ \hline
\includegraphics[height=.8em]{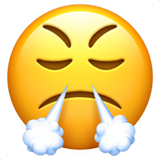} & 0.845\\ \hline
thats & 0.844\\ \hline
name & 0.826\\ \hline
life & 0.811\\ \hline
hate & 0.810\\ \hline
lazy & 0.803\\ \hline
\end{tabular}
\caption{Unigrams and their corresponding weights sorted in decreasing weight order (continued)}
\end{table}

\begin{table}[h]
\centering
\begin{tabular}{|c|c|} 
\hline
Unigram & Corresponding Weight\\ \hline
much & 0.799\\ \hline
mino & 0.791\\ \hline
!url & 0.788\\ \hline
!” & 0.785\\ \hline
knew & 0.747\\ \hline
tire & 0.698\\ \hline
bro & 0.691\\ \hline
dnt & 0.688\\ \hline
\includegraphics[height=.8em]{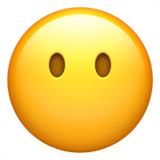} & 0.723\\ \hline
$ae\_$ & 0.687\\ \hline
\includegraphics[height=.8em]{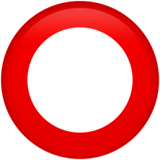} & 0.678\\ \hline
\$ & 0.658\\ \hline
wid & 0.657\\ \hline
pictures & 0.654\\ \hline
game & 0.651\\ \hline
\textbf{ever} & 0.642\\ \hline
st & 0.635\\ \hline
n***a & 0.625\\ \hline
world & 0.616\\ \hline
went & 0.616\\ \hline
who & 0.614\\ \hline
tw & 0.613\\ \hline
\includegraphics[height=.8em]{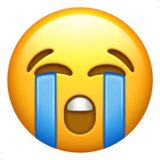} & 0.601\\ \hline
anotha & 0.578\\ \hline
\textbf{a} & 0.571\\ \hline
\includegraphics[height=.8em]{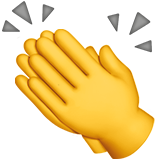} & 0.565\\ \hline
frm & 0.563\\ \hline
die & 0.562\\ \hline
mufuka & 0.535\\ \hline
irrelevant & 0.525\\ \hline
\includegraphics[height=.8em]{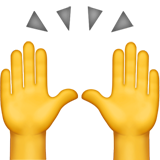} & 0.523\\ \hline
yall & 0.512\\ \hline
bet & 0.511\\ \hline
\textbf{out} & 0.508\\ \hline
snoop & 0.491\\ \hline
betta & 0.491\\ \hline
girl & 0.480\\ \hline
@user & 0.478\\ \hline
\textbf{that} & 0.460\\ \hline
\includegraphics[height=.8em]{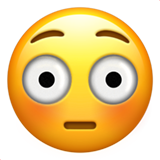} & 0.458\\ \hline
ain't & 0.447\\ \hline
watch & 0.434\\ \hline
torrance & 0.424\\ \hline
beef & 0.415\\ \hline
mouth & 0.407\\ \hline
mom & 0.406\\ \hline
\includegraphics[height=.8em]{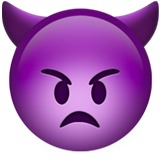} & 0.400\\ \hline
\end{tabular}
\caption{Unigrams and their corresponding weights sorted in decreasing weight order (continued)}
\end{table}

\begin{table}[h]
\centering
\begin{tabular}{|c|c|} 
\hline 
Unigram & Corresponding Weight\\ \hline
o'hare & 0.394\\ \hline
he & 0.391\\ \hline
's & 0.388\\ \hline
yo & 0.387\\ \hline
\includegraphics[height=.8em]{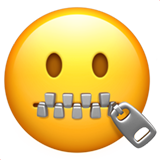} & 0.385\\ \hline
forever & 0.381\\ \hline
\#nolacking & 0.375\\ \hline
\includegraphics[height=.8em]{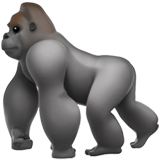} & 0.373\\ \hline
snitched & 0.360\\ \hline
gave & 0.355\\ \hline
if & 0.354\\ \hline
talk & 0.352\\ \hline
keep & 0.329\\ \hline
as & 0.327\\ \hline
internet & 0.325\\ \hline
\textbf{da} & 0.323\\ \hline
producer & 0.323\\ \hline
gang & 0.322\\ \hline
opp & 0.314\\ \hline
chica & 0.301\\ \hline
homie & 0.297\\ \hline
onna & 0.289\\ \hline
beat & 0.288\\ \hline
bitch & 0.287\\ \hline
\includegraphics[height=.8em]{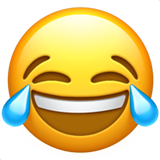} & 0.283\\ \hline
casper & 0.283\\ \hline
\#loony & 0.269\\ \hline
hoe & 0.260\\ \hline
whoever & 0.255\\ \hline
shootin & 0.252\\ \hline
\#mob & 0.197\\ \hline
lady & 0.183\\ \hline
rose & 0.183\\ \hline
fake & 0.181\\ \hline
\textbf{we} & 0.178\\ \hline
gun & 0.177\\ \hline
\textbf{be} & 0.176\\ \hline
smoked & 0.175\\ \hline
\#8tre & 0.170\\ \hline
caught & 0.162\\ \hline
jarocity & 0.155\\ \hline
! & 0.153\\ \hline
\includegraphics[height=.8em]{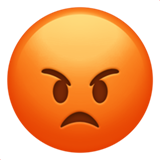} & 0.137\\ \hline
shordty & 0.126\\ \hline
just & 0.104\\ \hline
everyday & 0.090\\ \hline
nail & 0.076\\ \hline
\end{tabular}
\caption{Unigrams and their corresponding weights sorted in decreasing weight order (continued)}
\end{table}

\begin{table}[h]
\centering
\begin{tabular}{|c|c|} 
\hline 
Unigram & Corresponding Weight\\ \hline
snitch & 0.073\\ \hline
bumming & 0.071\\ \hline
\textbf{of} & 0.070\\ \hline
ona & 0.067\\ \hline
\textbf{u} & 0.058\\ \hline
steady & 0.052\\ \hline
like & 0.050\\ \hline
talm & 0.048\\ \hline
ways & 0.047\\ \hline
\#wuggaworld & 0.044\\ \hline
soft & 0.042\\ \hline
\#rt & 0.041\\ \hline
yellin & 0.041\\ \hline
\textbf{do} & 0.037\\ \hline
\textbf{in} & 0.035\\ \hline
\#cmb & 0.034\\ \hline
\#cvg & 0.029\\ \hline
\includegraphics[height=.8em]{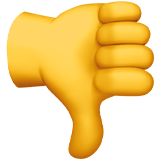} & 0.027\\ \hline
no & 0.025\\ \hline
\#051ym & 0.025\\ \hline
tutugang & 0.018\\ \hline
\#6775 & 0.010\\ \hline
\includegraphics[height=.8em]{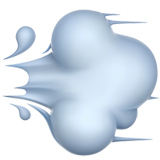} & 0.008\\ \hline
\#jar & 0.005\\ \hline
goof & 0.005\\ \hline
dime & 0.003\\ \hline
\#bricksquad & 0.003\\ \hline
\includegraphics[height=.8em]{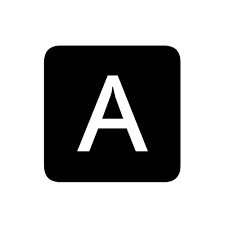} & 0.001\\ \hline
\end{tabular}
\caption{Unigrams and their corresponding weights sorted in decreasing weight order (continued)}
\end{table}

\begin{table*}[t]
\centering
\begin{tabular}{|c|c|c|c|c|c|c|}
\hline
\backslashbox{Models}{Unigrams} & in & be & do & any & u & out\\
\hline
\citet{blevins2016automatically} & 3.68 & 0.0$^{*}$ & 0.0$^{*}$ & 39.92$^{!}$ & 48.08$^{!}$ & 25.64$^{!}$\\
\hline
\citet{Chang2018DetectingGE} & 7.56$^{!}$ & 6.56$^{!}$ & 0.4 & 9.2 & 6.84$^{*}$ & 3.6\\
\hline
CNN + Twitter & 1.88$^{*}$ & 1.48 & 0.84 & 1.16$^{*}$ & 13.88 & 1.8\\
\hline
LSTM & 4.24 & 2.72 & 1.08$^{!}$ & 8.36 & 16.12 & 2.48\\
\hline
LSTM + Rationale & 3.96 & 1.12 & 0.08 & 3.64 & 12.76 & 1.4$^{*}$\\
\hline
\end{tabular}

\begin{tabular}{|c|c|c|c|c|c|c|}
\hline
\backslashbox{Models}{Unigrams} & basically & yea & ever & \includegraphics[height=.8em]{img/up_exclamation.png} & \includegraphics[height=.8em]{img/pointing_left.png} & we\\
\hline
\citet{blevins2016automatically} & 206.08$^{!}$ & 67.84$^{!}$ & 47.64$^{!}$ & 106.36$^{!}$ & 49.84$^{!}$ & 69.6$^{!}$\\
\hline
\citet{Chang2018DetectingGE} & 3.64$^{*}$ & 2.96 & 2.04 & 3.56 & 2.28$^{*}$ & 16.04\\
\hline
CNN + Twitter & 5.4 & 1.52$^{*}$ & 1.04 & 3.08 & 3.16 & 10.52$^{*}$\\
\hline
LSTM & 7.44 & 13.6 & 2.28 & 1.64 & 9.68 & 16.92\\
\hline
LSTM + Rationale & 4.16 & 8.0 & 0.16$^{*}$ & 0.48$^{*}$ & 3.92 & 11.96\\
\hline
\end{tabular}

\caption {The number of model's prediction flip from non-aggressive to aggressive, out of the 800 attacking tweets generated by inserting a specified ``neutral" unigram. 
To obtain a stable estimate, the count of number of flips of each model is averaged across 5 model runs and models trained on 5 folds - thus leading to non-integer results.
For each column, the worst performing entry is marked with ``!" and best with ``*". 
}
\label{additional_adversarial_stats}
\end{table*}